# On the Transferability of Learning Models for Semantic Segmentation for Remote Sensing Data

Rongjun Qin, *Senior Member, IEEE*, Guixiang Zhang, Yang Tang

*Abstract*— **Recent deep learning-based methods outperform traditional learning methods on remote sensing (RS) semantic segmentation/classification tasks. However, they require large training datasets and are generally known for lack of transferability due to the highly disparate RS image content across different geographical regions. Yet, there is no comprehensive analysis of their transferability, i.e., to which extent a model trained on a source domain can be readily applicable to a target domain. Therefore, in this paper, we aim to investigate the raw transferability of traditional and deep learning (DL) models, as well as the effectiveness of domain adaptation (DA) approaches in enhancing the transferability of the DL models (adapted transferability). By utilizing four highly diverse RS datasets, we train six models with and without three DA approaches to analyze their transferability between these datasets quantitatively. Furthermore, we developed a straightforward method to quantify the transferability of a model using the spectral indices as a medium and have demonstrated its effectiveness in evaluating the model transferability at the target domain when the labels are unavailable. Our experiments yield several generally important yet not well-reported observations regarding the raw and adapted transferability. Moreover, our proposed label-free transferability assessment method is validated to be better than posterior model confidence. The findings can guide the future development of generalized RS learning models. The trained models are released under this link: https://github.com/GDAOSU/Transferability-Remote-Sensing.**

## I. INTRODUCTION

SEMANTIC segmentation of remote sensing (RS) images has been a hot topic for the last two decades, as it presents one of the most fundamental tasks for various RS-based applications such as urban planning, environment monitoring, mining, mapping, and GIS (Geographical Information System) data updating [1]. Challenges may arise from the highly disparate data sources, complex scenes, level of uncertainties in spectrum and radiometry, atmospheric effects on imaging, training, and testing imbalances, as well as the ever-improving resolutions and sensors, etc. [2], leading to very poor model generalization and transferability when a trained model from one dataset is applied to another. Under this context, the transferability is measured as the achievable

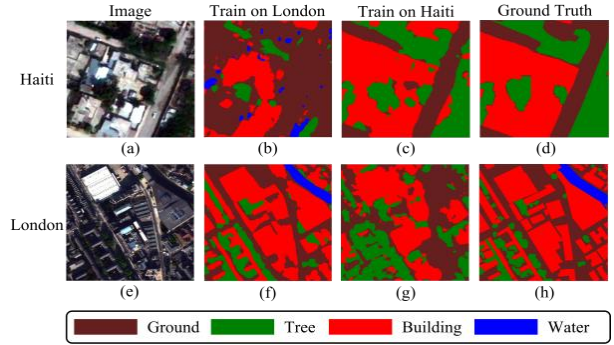

**Fig. 1.** An example of typical transferability issues. (a-d) present a sample in a dataset collected over Haiti and predicted results from a model (using Feature Pyramid Network) trained using another dataset collected over London (e-h), and vice versa. Models trained from a different dataset significantly underperform models trained from the same dataset.

accuracy when a trained model using one dataset is applied to an unseen dataset. This definition is slightly different from model generalization capability as defined in the learning theory [3], that the model generalization is the replicability of model performance when applied to unseen datasets as if they were performed in training (source) datasets. It is generally recognized that semantic segmentation for well-studied computer vision tasks, such as for automated driving and object segmentation, has shown to be a success. Once a model has been trained using sufficient data [4], it may be readily deployable for real-world applications for similar scenes. The success is primarily attributed to the well-structured scene for images taken at the ground level in a city area: sky on the upper part of the images, road/streets on the lower part, and facades of man-made structures on the left/right side of the image. However, the scene complexity in RS images is often underestimated: the content of RS images can be highly diverse, for example, residential and small buildings mixed with high-rise skyscrapers, large-sized commercial buildings, rivers, forests, arable lands, parks, etc. Despite newer models with novel deep learning (DL) architectures continuing to elevate new benchmark accuracies in baselines [5], the conclusions yielded from these works may not cast full confidence to end-users until these methods are practiced on users' data in their own problem set [6]. An example is shown in Fig. 1, where

The study is partially supported by the ONR grant (Award No. N000142012141 & N000142312670). (Corresponding author: Rongjun Qin, qin.324@osu.edu).

Rongjun Qin is with the Geospatial Data Analytics Lab, Department of Civil, Environment and Geodetic Engineering and the Department of Electrical and Computer Engineering, and Translational Data Analytics Institute, The Ohio state University, Columbus, USA (email: qin.324@osu.edu)

Guixiang Zhang is with the Geospatial Data Analytics Lab and Department of Electrical and Computer Engineering, The Ohio State University, Columbus, USA (email: zhang.10451@osu.edu)

Yang Tang is with the Geospatial Data Analytics Lab and Department of Civil, Environment and Geodetic Engineering, The Ohio State University, Columbus, USA (email: tang.1693@osu.edu)

datasets with completely different urban morphology (London vs. Haiti) challenge the transferability of DL models.

Existing literature regards such data differences as domain gaps, and approaches to minimizing these gaps are called domain adaptation (DA), which variably assume these gaps be from known causes (e.g., lack of radiometric calibration) [7], some of known causes [8], or completely unknown [9]. The terminology used in this context defines the training dataset as the source domain (datasets with labels) and the test data (dataset with little or no labels) as the target domain. These approaches may assume the availability of a few labels in the target domain, which is then called semi-supervised DA or none (i.e., unsupervised DA). There have been promising progresses in DA approaches to aid the transferability of models [9]; however, most of these studies work on closed and sometimes synthetic data [10], yet there is a general lack of understanding of the raw transferability of basic learning models, as well as how much different DA algorithms can improve. Here, the *raw transferability* of the model refers to its predicting accuracy towards different and unseen datasets. When a DA method is applied to the model, the transferability of this model is defined as *adapted transferability*. As of now, there are two fundamental gaps in this line of research: 1) There is no systematic study that quantifies model transferability and evaluates the effectiveness of DA in RS semantic segmentation; 2) There are no label-free measures to predict how much domain gaps are when a model is applied to a test data.

In this paper, we aim to investigate these two research questions by performing a transferability study on different learning models. This is performed by leveraging the use of ever-growing public benchmark datasets (including our data), as well as testing different DA methods on their ability to improve the transferability. This includes five commonly used deep learning (DL) models, namely, U-Net [11], DeepLabv3+ [12], DeepLabv3+ with attention module (Attention) [13], Feature Pyramid Network (FPN) [14], and HRNet [15], and one traditional model (Random Forest [16]) as the base models. Three different DA methods, including AdaptSegNet [9], Category-Level Adversarial Network (CLAN) [17], and ScaleAware [13], are evaluated in this study. Those models are cross tested through large datasets with varying sources (sensors), geographical regions, and scenes, with and without height information. The testing accuracy is quantified to understand the raw and adapted transferability of these models.

In addition, we develop a paradigm to predict the domain gaps following extremely simple heuristics: we assume that the well-practiced and spectral RS indices (e.g., NDVI (normalized difference vegetation index [18]) as a medium to evaluate domain gaps: we found that evaluating a model based on simple classes derived from these indices may provide good reference on the model's transferability. We expect that the conclusions and statistics drawn from this work will provide a baseline of the model transferability to improve the basic understanding of domain gaps and the predicting metrics, thus guiding RS DA approaches to be more robust and measurable.

The remainder of this paper is organized as follows: Section II. introduces domain gaps discussed by existing works; Section III. Dataset And presents details of the dataset used in our work and describes experiment setups. Section IV. presents the analysis of the results; Section V presents our study on our proposed method to measure model transferability using RS indices; Section VI. concludes our analysis.

## II. RELATED WORK

The boost of deep models has been consistently updating the benchmark performances of well-defined machine learning tasks. In these tasks, the training data captures the distribution of the testing data well [19]. However, poor model performances were often attributed to the lack of appropriate training data for capturing the distribution of the target datasets [20]. Globally, the level of data variation in terms of sensors, atmospheric conditions, and scene contents are extremely diverse such that, in [21]'s work, they noted every single RS image to be a single domain generally requiring DA approaches to accommodate, similar statements were found in a recent Science paper outlining the challenges about domain gaps in RS tasks [22]. Thus, the community mostly put forth efforts into 1) developing more public benchmark datasets and 2) developing DA and transfer learning approaches to accommodate the presumed domain gaps.

### A. Public Benchmark Datasets

The benchmark dataset culture is most popular in the computer vision community, and the success of DL models as a revolution is through the ImageNet benchmark dataset, with millions of images to feed into a network model with a large capacity [23]. The DL campaigns start with the datasets to develop models with higher accuracy and better generalization for recognition tasks in everyday objects. It was noted in [24] that an RS version of ImageNet is desired; although efforts of such datasets started almost 15 years ago from the IEEE data fusion contest [25], it has recently seen growth from 2015-2021 through a few public competitions such as IARPA functional map of the world [26], SpaceNet [27], BigEarthNet [28], EarthNets [29] and Deep Globe competitions [30], which has accumulated approximately thousands of Gigabytes (GB) of image and annotated data. These datasets, despite their volume, are still likely insufficient for constructing a one-for-all solution as compared to the diversity of scenes in RS tasks.

### B. Domain Adaptation

The DA is generally developed under the hypothesis that biases exist between the source (i.e., training) and the target (testing) domain in terms of their data and/or feature distribution, and these biases are correctable. Such a correction can be performed at the image level, feature level, or classifier level: the image-level DA aims to perform style transfer [31] through generative networks (or regression networks) to transfer the source image to have the same style of target domain images; DA at the feature-level assumes the biases between the source and target domain can be reduced by parametrically aligning the features extracted at different stages of the model through transformations or optimized network weights [32]; the classifier-level DA, rather than aligning the features between two domains, trains a shared classifier to effectively classify on both domains, such as manifold learning [33] or using domain-invariant discriminative features combine with distance-based regularization [34].

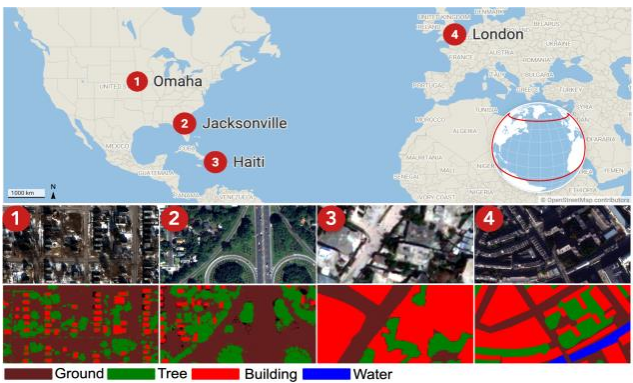

**Fig. 2.** An overview of the geographical distribution of datasets and their diverse urban morphology

When applied to RS data, these approaches assumed different types of biases such as radiometry, spectrum, or scale differences based on examples. It presents different levels of domain gaps; for example, the work of [35] regards neighboring tiles of images as "light" and far-away tiles as "large" domain gaps, and the work of [36] regards data from different times to possess more domain gaps, etc. These domain gaps can sometimes be measured through data distribution metrics, such as the Fréchet distance [37], rotation prediction accuracy [38], and the difference of confidences [39]. These metrics, on one hand, can be used as part of the goal functions to minimize DA, and on the other hand, can be used to estimate the level of data domain gaps for downstream applications such as adaptive model selection for model ensemble [40].

While most of the ongoing efforts focus on collecting more labeled datasets and novel DA approaches, there is a general lack of understanding of the model's raw and adapted transferability and ways to assess them when target labels are not available. To the authors' best knowledge, there is no comprehensive evaluation of these aspects; tests were often performed using one or two datasets covering only a very small region [41]. With the growing volume and diversity of individual public benchmark datasets, we consider that it is possible to cross-validate among them to study and quantify the transferability of learning models. In this paper, we perform this experimental analysis by using a few public and our own datasets following a simple rationale: we regard datasets of diverse geographical regions as different domains, train models from one domain and test them on the other using different models. We quantify the transferability of these models and develop metrics to predict the transferability in occasions where ground truth is not available in the target domain.

## III. DATASET AND EXPERIMENT SETUP

### A. Experiment Datasets

As mentioned above, our experiments evaluate the transferability of models using four datasets, two publicly released datasets and two of our own, with significant differences in volume, satellite sensors, resolution, geographical location, and scenes among different datasets. As shown in Fig.2. and TABLE I. Each dataset has images with multispectral (MSI) bands. Semantic labels and digital surface model (DSM) or above ground level (AGL) as height information (through stereo-photogrammetry or LiDAR) are also available. The use of MSI and height information allows us to train models with multi-modal information to understand their contribution/impact to model transferability. More details of these datasets can be found in TABLE I. The volume of each dataset is measured as the number of patches sampled to 0.31m resolution to match the training data; each patch has a size of 1024×1024 pixels. Among all the datasets listed in TABLE I, JAX, OMA, and London datasets have sufficient data to train deep networks, and their scene morphology is distinct enough. Therefore, our experiments mostly train deep networks on the JAX, OMA, or London dataset (as source domains) and apply the trained model to the rest (as target domains). This gives us the possibility to analyze models' performances when they were trained by different datasets.

TABLE I
DATASETS USED IN THIS STUDY

| Dataset | # Patches | Bands | Sensor | GSD | Region | Scene |
|---------|-----------|-------|--------|-----|--------|-------|
| JAX | 1015 | 8 | WV-3 | 0.31m | US | U |
| OMA | 1768 | 8 | WV-3 | 0.31m | US | U |
| Haiti | 144 | 4 | GE-1 | 1.84m | US | DU |
| London | 529 | 4 | WV-3 | 0.50m | EU | DU |

U: Urban; **DU**: Dense Urban; **WV**: Worldview; **GSD**: Ground Sampling Distance.

**JAX, OMA**: These datasets are from the 2019 Data Fusion Contest [25], which provides Urban Semantic 3D (US3D) data, a large-scale public dataset including multi-view, multi-band satellite images and ground truth geometric and semantic labels for two big cities, Jacksonville (JAX), Florida and Omaha (OMA), Nebraska, U.S.

**Haiti, London:** These two datasets are collected to further diversify the geographical region. Both the London and Haiti datasets are from different continents with different land/urban morphology. Specifically, the Haiti dataset can be particularly challenging due to its unique Middle East urban morphology. Both datasets have stereo/multi-stereo images to be further processed to derive digital surface models and orthophotos. The semantics of the datasets are manually annotated.

### B. Experiment Setup

One traditional method, i.e., Random Forest (RF), and five deep learning (DL) networks, i.e., U-Net [11], Deeplabv3+ [12], Deeplabv3+ with attention module [13], FPN [14], and HRNet [15] are adopted as base models in our experiment. In addition, three DA approaches, i.e., AdaptSegNet [9], CLAN [17], and ScaleAware [13] are employed for our evaluation. Models with different setups were trained, which include base models trained using RGB bands of the image, models trained using both the multispectral bands (RGB and Near-infrared (NIR), denoted as RGBN), and height information (i.e., multi-modal model), as well as adapted models to which the DA methods were applied. It should be noted that these base models were originally designed for RGB images, and to allow multi-modal input (RGBN+height), we changed the number of the channels of the first convolutional layer for DL models, and the number of feature bands used for RF. Thus, the size of the model parameters may be appropriately prorated. More information is listed in TABLE II. and the suffix "_H" means that in multi-modal models where height information is included as the input. Different base models are listed in terms of their complexity (number of trainable parameters), where Deeplabv3+ is the lightest (12.3M) and HRNet (65.9M) is the heaviest. Since RF does not have a fixed parameter structure (depending on data/feature), thus it is listed as a separate category, but it can be generally regarded as a light/shallow classifier than all DL ones. TABLE II also listed the hyperparameters used for model training, and more details of data preparation and process are introduced in the subsections below.

TABLE II
TESTED MODELS AND HYPERPARAMETERS

| Models | Input | # of parameters | DA approaches | Hyperparameters (base model) |
|---|---|---|---|---|
| Deeplabv3+ (_H) | R, G, B, NIR (, AGL) | 12.3M | | **Batch size**: 2; **Initial learning rate**: 0.02 |
| Attention (_H) | R, G, B, NIR (, AGL) | 12.3M | AdaptSegNet, | **Learning rate decay power**: 0.9 |
| FPN (_H) | R, G, B, NIR (, AGL) | 13.1M | CLAN, | **Momentum**: 0.9 |
| U-Net (_H) | R, G, B, NIR (, AGL) | 14.3M | ScaleAware | **Weight decay**: 0.0001 |
| HRNet (_H) | R, G, B, NIR (, AGL) | 65.9M | | **Optimizer**: Stochastic Gradient Descent (SGD) |
| RF (_H) | R, G, B, NIR, GLCM (, AGL) | \ | \ | **Number of trees**: 500; **Max depth**: 20<br>**Min samples leaf**: 1000; **Min samples split**: 4000<br>**Number of samples**: 4,000,000; **GLCM window**: 13×13 |

**Label Preparation**: The annotation protocols of different datasets from their original sources are different. For example, they may have different numbers of classes and different class definitions at different granularity. To perform a fair analysis, we covert these labels to four major and consistent classes: ground, tree, building, and water. This process can be achieved effectively by building a lookup table per dataset, by mostly discarding small classes and merging fine-grained classes into coarse ones. For example, the JAX and OMA datasets originally had five classes: ground, tree, roof, water, and elevated road. We convert elevated road class to "void". For the Haiti and London datasets, the road, impervious, agriculture, grassland, and barren classes are all merged into the ground, and the shrubland class is changed to void. Other classes, including tree, building, and water classes remain unchanged. The class of "void" will not be used for training or evaluation.

**Data Pre-processing**: For all bands (R, G, B, and NIR), we normalized the original 16-bit pixel value to 0 ~ 1 based on histogram truncation of the entire dataset (cutting upper and lower 2%). JAX and OMA datasets come with AGL derived from LiDAR. For the other two datasets (stereo/multi-stereo), we first apply the RPC stereo processor (RSP) software [42], [43] to derive the digital surface models (DSM) and orthophotos at the resolution of 0.31m. The DSM was further processed to height maps (i.e., AGL) using a morphology-based building extraction algorithms [44]. The orthophotos and DSMs are then prepared into fixed-sized patches for model training. Pixel/AGL values are rescaled to 0 ~ 1 following the same histogram truncation approach. Each image of these datasets in training sets is prepared as 512 × 512 patches with an overlap ratio of 0.5. The same process is performed on the testing patches to allow the same-sized input for inference. The use of this overlap is to facilitate a more robust label prediction, which we use to aggregate probabilities of classes' overall repeated pixels in the overlapping region to vote for the predicted label.

**Base Models**: Five DL models, three DA approaches and the RF are implemented in this experiment as below:
a) U-Net [11], DeepLabv3+ [12], FPN [14]: These three networks are among the most commonly used models in the RS semantic segmentation. These models come with variants depending on the encoder. In our experiment, to make a fair comparison, we use a ResNet of 18 layers as the backbone encoder for all these models.

b) DeepLabv3+ with attention module (Attention) [13]: DeepLabv3+ is also a widely used network structure. The original network has an Atrous Spatial Pyramid Pooling (ASPP) module that may fall short of providing high resolution output. The recent work [13] suggested that adding a self-attention module can enhance the learned feature maps and enhance the performance.

c) HRNet [15]: In our study, we employ the HRNetV2 model, An enhanced version of the popular HRNet [45] that is stressed for predicting high resolution labels. It fuses multi-resolution branches at all times to the main details. HRNetV2 enhanced the original network by upsampling the low-resolution layers to higher ones prior to fusion. Since this is a minor modification, for simplicity, we will refer to HRNetV2 as 'HRNet' throughout this paper.

**Domain Adaptation Methods**: Three commonly used DA methods are evaluated, which include AdaptSegNet [9], ScaleAware [13], and CLAN [17]. The general concept of these DA approaches is to adopt adversarial learning to the semantic networks to achieve better aligned feature maps. We adopted the original codes released by the authors of these approaches in our experiment. It should be noted that all DA methods in our evaluation do not require target labels, which therefore, belong to unsupervised DA.
a) AdaptSegNet [9]: This approach trains a discriminator in the output space (semantic maps). The idea is to "discriminate" semantic maps that do not resemble those produced from the "source" data. Through positive/negative samples, the network trains to extract features that are less invariant.

b) ScaleAware [13]: We denote this DA approach as "ScaleAware" since this approach was proposed to address the scale difference of scenes among RS imageries. In addition to a standard feature discriminator in the output space like AdaptSegNet, this approach adopts another scale discriminator to discriminate scale differences among images.

c) CLAN [17]: The Category-level Adversarial network (CLAN) is an enhanced version of AdaptSegNet but follows a different feature alignment scheme, as it tries to perform local feature alignment by weighting per-pixel adversarial losses based on a precision-based prediction. Results in its original paper show a better performance than typical DA methods that only perform global alignment in the feature space.

d) Random Forest [16]: The random forest (RF) is trained to perform pixel-wise classification for each RS image, 500 trees are used in the model. In addition to the original 4-band

TABLE III
mIoU OF BASE MODELS AND THEIR RAW TRANSFERABILITY

| | Without Height Information | | | | | | | | | | | | | | Haiti |
|---|---|---|---|---|---|---|---|---|---|---|---|---|---|---|---|
| **Source** | **JAX** | | | | | **London** | | | | | **OMA** | | | | **Haiti** |
| **Target** | JAX (S2S / T2T) | London (S2T) | OMA (S2T) | Haiti (S2T) | Average (target) | London (S2S / T2T) | JAX (S2T) | OMA (S2T) | Haiti (S2T) | Average (target) | OMA (S2S / T2T) | JAX (S2T) | London (S2T) | Average (target) | Haiti (T2T) |
| RF | 48.04 | 27.22 | 27.90 | 22.57 | 25.90 | 53.10 | 26.21 | 20.06 | 31.41 | 25.89 | 31.69 | 26.74 | 26.21 | 26.48 | 35.44 |
| Deeplabv3+ | 84.16 | 36.53 | **47.70** | 24.25 | **36.16** | 67.29 | 41.07 | 32.94 | 25.00 | 33.00 | 68.35 | 57.61 | 40.98 | 49.30 | 49.66 |
| Attention | 84.78 | 28.55 | 39.95 | 18.89 | 29.13 | 67.52 | **46.62** | **35.33** | 25.67 | 35.87 | 74.99 | 62.41 | **49.71** | **56.06** | **49.80** |
| FPN | 85.70 | 31.35 | 41.74 | 21.92 | 31.67 | 67.68 | 44.67 | 33.71 | 27.64 | 35.34 | **74.87** | **64.15** | 45.53 | 54.84 | 49.55 |
| U-Net | **85.97** | 33.22 | 47.13 | 22.34 | 34.23 | **67.98** | 45.42 | 34.35 | 24.97 | 34.91 | 74.60 | 61.46 | 38.50 | 49.98 | 48.34 |
| HRNet | 75.28 | **37.19** | 43.32 | **24.47** | 34.99 | 66.63 | 45.12 | 31.59 | **31.45** | **36.05** | 65.28 | 57.03 | 41.58 | 49.31 | 45.76 |
| Average (DL) | 83.18 | 33.37 | 43.97 | 22.37 | 33.24 | 67.42 | 44.58 | 33.58 | 26.95 | 35.04 | 71.62 | 60.53 | 43.26 | 51.89 | 48.62 |
| | **With Height Information** | | | | | | | | | | | | | | |
| RF_H | 74.83 | 35.25 | 49.92 | 28.28 | 37.82 | 56.88 | 41.86 | 38.24 | 34.21 | 38.10 | 62.43 | 57.49 | 35.35 | 46.42 | 38.86 |
| Deeplabv3+_H | 89.64 | 39.48 | 66.28 | 27.39 | 44.38 | **67.42** | 50.65 | 37.37 | 31.46 | 39.83 | 77.19 | 74.06 | **46.53** | **60.30** | **49.74** |
| Attention_H | 86.38 | 38.75 | 57.48 | 26.72 | 40.98 | 67.00 | **55.86** | **42.26** | 33.97 | **44.03** | 80.93 | 75.16 | 44.45 | 59.81 | 49.48 |
| FPN_H | 90.07 | 35.82 | 64.67 | 28.29 | 42.93 | 67.25 | 42.88 | 35.57 | 32.08 | 36.84 | **84.13** | 73.71 | 42.33 | 58.02 | 49.03 |
| U-Net_H | **91.40** | **42.12** | **66.34** | **29.94** | **46.13** | 67.13 | 47.59 | 35.72 | **34.43** | 39.25 | 80.94 | 74.28 | 45.16 | 59.72 | 48.72 |
| HRNet_H | 79.96 | 36.22 | 59.74 | 26.68 | 40.88 | 66.73 | 45.98 | 34.97 | 30.65 | 37.20 | 80.71 | **76.14** | 40.31 | 58.23 | 49.47 |
| Average (DL) | 87.49 | 38.48 | 62.90 | 27.80 | 43.06 | 67.15 | 48.59 | 37.18 | 32.52 | 39.43 | 80.78 | 74.67 | 43.76 | 59.21 | 49.29 |

**Fig. 3.** An example of visual results on base models applied to different datasets. Results of the best performing base models according to TABLE III are selected in this Figure.

multi-spectral information, we incorporate the gray-level co-occurrence matrix (GLCM) [46], a 6-dimension texture feature vector for each pixel into the feature vector. The feature dimension of each pixel is 10 and 11 for RF and RF_H, respectively.

**Model Hyperparameters:** Most of the hyperparameters used in the experiments are taken based on the respective original papers. For RF, we tune these parameters to get the best empirical performance. More details are in TABLE II. The hyperparameters in the table are applied to all DL models.

**Evaluation Metric:** The results are evaluated by the mean intersection of union (mIoU):

$$mIoU = \frac{1}{N_c} \sum_i \frac{\# \ pixel \ of \ (C_i^{prediction} \ \cap \ C_i^{ground \ truth})}{\# \ pixel \ of \ (C_i^{prediction} \ \cup \ C_i^{ground \ truth})} \quad (1)$$

where $C_i^{prediction}$ denotes pixels labeled as the i[th] class in the model prediction and $C_i^{ground \ truth}$ denotes pixels labeled as the i[th] class in the ground truth. $N_c$ (=4 in our experiment) is the total number of classes. In addition, for each pair of domains and models, confusion matrices are generated for evaluation.

**Evaluation Protocol:** Considering that each dataset presents a domain, we split each dataset into two portions, a training set and a validation set, with a ratio of 8:2. For training the base models (without DA) approaches, we train the models on the training set of the source domain using a pocket algorithm [3]: we save the checkpoint every 5k iterations. After 100k iterations, we validate each checkpoint in the validation set of the target domain and select the checkpoint with the highest mIoU as the raw transferability of this model to this target domain. For training adapted models (with DA), we use the entire labeled source domain and unlabeled training set of the target domain for training. The models are validated on the validation set of the target domain in the same way to get the adapted transferability. This evaluation protocol takes the best performing checkpoint and keeps the validation set target untouched in the training process. We evaluate the transferability of models under three configurations: source-to-source (S2S), source-to-target (S2T), and target-to-target (T2T). S2S or T2T represents train and testing on the same source (JAX or London) or target domain (OMA and Haiti). S2T means training and validating models on the source domain and testing on the target domain to evaluate the transferability.

## IV. RESULTS AND ANALYSIS

TABLE III. and TABLE IV. listed the results following the experiments of base model performance (raw transferability) and DA approach performance (adapted transferability) in Section III-B. Table III shows the results of the traditional method RF and five DL segmentation models without DA approaches. The source and target datasets are in the first two columns. The mIoU of different models is evaluated based on their ground-truth labels using Equation (1). JAX, OMA and

TABLE IV
S2T mIOU of Models with DA Approaches

| Data | Source | Target | Deeplabv3+ | | | | Attention | | | | FPN | | | | U-Net | | | | HRNet | | | | Average | | | |
|---|---|---|---|---|---|---|---|---|---|---|---|---|---|---|---|---|---|---|---|---|---|---|---|---|---|---|---|
| | | | R | AS | CL | SA | R | AS | CL | SA | R | AS | CL | SA | R | AS | CL | SA | R | AS | CL | SA | R | AS | CL | SA |
| Without Height Information | Jax | London | 36.5 | 38.6 | 52.6 | 51.0 | 28.6 | 53.9 | **54.3** | 56.7 | 31.4 | **54.3** | 48.6 | 53.3 | 33.2 | 50.4 | 51.8 | 53.0 | 37.2 | 50.6 | 46.1 | 52.4 | 33.4 | 49.4 | 50.7 | 53.3 |
| | | OMA | **47.7** | 48.8 | 49.2 | 50.3 | 40.0 | 43.0 | 43.2 | 45.1 | 41.7 | 49.9 | 48.1 | **54.6** | 47.1 | **51.8** | **49.5** | 53.3 | 43.3 | 46.4 | 38.9 | 43.7 | 44.0 | 48.0 | 45.8 | **49.4** |
| | | Haiti | 24.3 | 27.2 | 27.0 | 29.0 | 18.9 | 27.4 | 24.1 | 29.1 | 21.9 | 28.0 | 25.3 | **30.5** | **24.5** | **30.9** | 27.3 | 27.8 | 22.4 | 27.5 | 22.8 | 27.1 | 22.4 | 28.2 | 25.3 | **28.7** |
| | | Average | **36.2** | 38.2 | **42.9** | 43.4 | 29.1 | 41.1 | 40.5 | 43.6 | 31.7 | **44.1** | 40.7 | **46.3** | 35.0 | 42.6 | 37.4 | 41.3 | 34.3 | 41.5 | 35.9 | 41.1 | 33.2 | 41.9 | 40.6 | **43.8** |
| | London | JAX | 41.1 | 48.8 | 50.6 | 49.6 | **46.6** | 50.1 | 50.2 | 50.8 | 44.7 | **51.2** | 50.7 | **53.1** | 45.4 | 48.6 | 47.8 | 51.5 | 45.1 | 50.5 | 47.7 | 49.4 | 44.6 | 49.8 | 49.4 | **50.9** |
| | | OMA | 32.9 | 38.5 | 38.0 | 39.1 | **35.3** | **40.6** | 38.3 | 41.7 | 33.7 | 40.3 | **39.0** | **42.8** | 34.4 | 40.1 | 34.7 | 40.8 | 31.6 | 38.6 | 36.1 | 37.6 | 33.6 | 39.6 | 37.2 | **40.4** |
| | | Haiti | 25.0 | 33.4 | 30.6 | 34.5 | 25.7 | 30.5 | 29.4 | 35.0 | 27.6 | 31.5 | **31.7** | 34.2 | 25.0 | 32.4 | 29.9 | 34.8 | **31.5** | **35.0** | 32.0 | **37.2** | 27.0 | 32.6 | 30.5 | **34.9** |
| | | Average | 33.0 | 40.2 | 39.7 | 41.1 | 35.9 | 40.4 | 39.3 | 42.5 | 35.3 | 41.0 | **40.5** | **43.4** | 34.9 | 40.4 | 37.4 | 42.4 | **36.1** | **41.4** | 38.3 | 41.0 | 35.0 | 40.7 | 39.0 | **42.1** |
| With Height Information | Jax | London | 39.5 | **51.1** | **48.2** | 46.5 | 38.8 | 44.0 | 46.6 | 47.3 | 35.8 | 42.8 | 36.2 | **48.2** | **42.1** | 44.4 | 45.4 | 45.5 | 36.2 | 41.4 | 42.0 | **49.7** | 38.5 | 44.7 | 43.7 | **47.4** |
| | | OMA | 66.3 | 66.3 | 68.1 | 69.0 | 57.5 | 61.7 | **70.0** | 63.9 | **64.7** | **67.5** | **64.7** | **70.5** | 66.3 | **68.6** | **69.6** | **71.6** | 59.7 | 63.0 | 62.6 | 58.7 | 62.9 | 65.4 | **67.0** | 66.7 |
| | | Haiti | 27.4 | 27.1 | 25.8 | 26.6 | 26.7 | **29.6** | 27.3 | 26.6 | 28.3 | 25.8 | **28.2** | 23.6 | **29.9** | 26.2 | 27.6 | 25.6 | 26.7 | 22.9 | 24.1 | 25.9 | **27.8** | 26.3 | 26.6 | 25.7 |
| | | Average | 44.4 | 48.2 | 47.4 | 47.3 | 41.0 | 45.1 | **48.0** | 46.0 | 42.9 | 45.4 | 43.1 | 47.4 | **46.1** | **46.4** | **47.5** | 47.6 | 40.9 | 42.4 | 42.9 | 44.8 | 43.1 | 45.5 | 45.8 | **46.6** |
| | London | JAX | 50.7 | **54.4** | 52.8 | 56.5 | **55.9** | 56.6 | 57.2 | 56.0 | 42.9 | 55.5 | 57.2 | 57.2 | 47.6 | 53.9 | **60.5** | **57.9** | 46.0 | 56.1 | 53.9 | 56.2 | 48.6 | 55.3 | 56.3 | 56.8 |
| | | OMA | 37.4 | **43.7** | 42.0 | 41.9 | **42.3** | 41.7 | 40.6 | **44.5** | 35.6 | 42.7 | **44.4** | 44.0 | 35.7 | 42.1 | 41.0 | 43.7 | 35.0 | 40.5 | 40.6 | 36.2 | 37.2 | 42.1 | 41.7 | **43.2** |
| | | Haiti | 31.5 | 35.0 | 33.8 | 37.7 | 34.0 | 36.4 | 36.2 | **37.9** | 32.1 | 35.6 | 36.0 | 37.3 | 34.2 | 36.4 | 32.0 | 37.2 | 32.5 | 35.9 | 34.8 | 37.1 | 32.8 | 35.8 | 34.1 | **37.2** |
| | | Average | 39.8 | 44.4 | 42.9 | 45.4 | 44.0 | **44.9** | 44.7 | 46.1 | 36.8 | 44.6 | **45.8** | 46.2 | 39.3 | 44.1 | 44.5 | **46.3** | 37.2 | 44.0 | 42.3 | 44.7 | 39.4 | 44.4 | 44.0 | **45.7** |

**R:** Raw; **AS:** With AdaptSegNet Approach; **CL:** With CLAN Approach; **SA:** With ScaleAware Approach.

London datasets were used to train different base models. Base models were also trained and tested on the Haiti dataset, the mIOU of which is used to indicate the best achievable accuracy for this dataset. Since the Haiti dataset is significantly small in volume, it is not used to test other datasets.

**Raw Transferability:** From the highlighted average mIoU of five based DL models in TABLE III, we observe that models under S2S and T2T configurations obviously have achieved much higher accuracy (over 30% better) than the models under the S2T configuration, which is expected, as we assume there are no domain gaps under S2S and T2T configurations. Although these S2S and T2T mIoU vary with the models and datasets (domains), they generally achieved mIOUs of over 67%. However, only 49% of mIOU is achieved in the Haiti dataset, which is mainly due to the small volume of the dataset and the very cluttered scene. On the other hand, the mIoU of RF also has different performances among different experiments: for example, as a shallow classifier, RF achieved 48.04% of mIOU when using only the image information (without height information) for the JAX dataset, while it is significantly better when the height information is used (74.83%), which aligns with conclusions from past studies on shallow classifiers using height information [47], [48]. Other than these general observations, we have a few observations regarding the raw transferability of these models:

First, the raw transferability of a model is domain specific. For example, from the highest mIOU (bolded) on each dataset, the performances of the model in the S2S scenario do not necessarily indicate their performance in the S2T scenario when the height information is not used. For example, the top-performing model in the S2S scenario of JAX is U-Net, while when it was applied to the other three datasets (London, OMA, Haiti), it was not the best performing model. However, a certain consistency is observed when the height information is used: U-Net_H achieves the best performance in both S2S and S2T scenarios for JAX dataset, while for London dataset, despite Attention_H is not the best performing model (Deeplabv3+_H is), it consistently achieves the best performance when applied to other datasets (high raw transferability), which outperforms Deeplabv3+_H to a notable level. There is more obvious evidence supporting the same claim: the RF classifier generally achieves significantly worse mIOU, while when it is transferred (applied) to the Haiti dataset, it ties with or is better than DL models. In addition, the best performing model in each scenario does not depend on the complexity of the model. Based on the learning theory, it is possible that the best performing model is the one that has a complexity close to the VC dimension of the problem (dataset + task) [3].

Second, the average S2T mIOU of DL models varies with the source domains they are trained from and shows no significant correlation with the data volume despite the fact that the London domain only has 529 patches while the JAX has 1,015 patches, when transferred to other datasets (without using height information), the models trained on the London domain outperform the ones trained on JAX domain (35.04% vs. 33.24%). Although, when involving the height information, the models trained on the JAX dataset outperform the models trained on the London domain in terms of average mIOU (43.06% vs. 39.43%). This higher average mIOU may be attributed to the fact that the OMA dataset has less domain gap to JAX (same sensor, same country, and same urban morphology). In addition, models trained on the London dataset are generally better than models trained on the JAX dataset when applied to the Haiti dataset. With educated guess, we attribute this to the fact that different domains might capture different data distribution; for example, a domain with a smaller number of patches, might have captured evenly distributed urban patterns, in comparison to the domain with many sample patches but densely captured similar urban patterns.

Third, models using height features (i.e., AGL) generally reflect better raw transferability in terms of the average S2T mIOUs. This is lightly correlated to our first observation, that models with height information have a certain consistency in S2S and S2T scenarios. In addition, with height information, we have also observed that more transferability improvement occurs at the shallow classifier RF in this experiment. For example, the average mIoU of RF trained on the London dataset and applied to the other target datasets increases from 25.89% to 38.10% (12.21% better), and DL models' mIoUs only increase from 35.04% to 39.43% (4.39% better on average). Note, that we exclude the statistics from OMA to JAX since these two domains are very similar. These advances can be mainly due to the nature of geometric/height information being less variant on different objects: for example, it is less impacted by illumination, sensors, and represents explicitly the physical dimensional information of the objects. The fact that the DL model has less improvement when incorporating height information, may be due to that these complex models may have implicitly captured the geometry information of the scene.

**Fig. 4.** Examples of results of best performing adapted models based on Table IV.

To provide a visual intuition, we have included the result for one out of a thousand patches, as shown in **Fig**. 3. Visual results of this patch generally agree with the statistics shown in TABLE III. Due to space limitation, for each sub-figure, we selected the best performing models according to TABLE III.

**Adapted Transferability:** In TABLE IV, three different DA approaches were implemented to evaluate the adapted transferability of five DL models under the S2T scenario. For readability, we have included the mIOU of base models in TABLE III into TABLE IV. We compared the performance of models with and without DA approaches. On average, adapted transferability is notably higher than the raw transferability of DL models in most cases, and among the three DA approaches, ScaleAware outperforms the other in terms of the transferred (S2T) mIOU, evidenced by the average mIOU in TABLE IV. When focusing specifically on each DA method, we did not find a consistent pattern that a certain DA approach works better over one base DL model or another. A DA approach may work best on one DL model in one dataset combination (source and target), while underperforming another DL model in another dataset combination. Moreover, we have also observed there are certain cases, where these DA approaches cannot improve the transferability of the model, for example, transferring from JAX to the Haiti dataset (with height information). With educated guess, we consider that this is due to that the T2T mIOU for Haiti is already very low, and height information itself may play a role in closing the adaptable domain gaps; thus, there is not much room for improvement. This observation is consistent, as the margin of improvement by DA is smaller when height information is used.

**Fig.** 4 shows some selected results before and after DA. Due to space limitations, for results with a certain DA method, we select the base model that produced the best results according to TABLE IV. The reader may find more visual examples under our supplemental material on GitHub.

## V. An Unsupervised Method For Model Transferability Prediction

### A. Concept

The analysis in Section IV concludes that the transferability of a trained model is highly variable with respect to the domains and inputs. While DA approaches help, the improvement is still marginal as compared to the T2T scenario. Therefore, the raw transferability still generally decides the usability of a trained model when being applied to the target dataset. However, evaluating the transferability of a trained model on a target domain (with no labels) is an ill-posed problem, as there exists no reference to the ground truth of the predicted labels. Remote sensing image interpretation has existed for decades, and there have been empirical indices that utilize multispectral information. Examples of such indices are normalized differences in the vegetation index (NDVI) [18], normalized differences in the water index (NDWI) [49], etc., which are linearly correlated with the likelihood of certain land-cover classes. These indices are robust to many disturbing factors, such as illumination and sensor responses [50]. Although land-cover classes predicted by these indices cannot compete with the supervised method, they have demonstrated relatively good generalization capability using general multispectral RS images, evidenced by their heavy use in numerous studies in RS for crops, forests, and water [51]. Therefore, we hypothesize that these indices might be attempted as valid cues for examining the transferability of models on target domains, with a very simple and heuristic idea: We consider using land-cover class labels predicted by these indices as the pseudo ground truth label to evaluate a model, with the expectation that good performances of models evaluated by these indices generally show good model transferability.

The underlying rationale of this idea can be loosely explained as follows: denote $\mathcal{P}_{sup}(Y|X)$ and $\mathcal{P}_{index}(\widehat{Y}|X)$, respectively as the estimated probabilistic distribution of a supervised model (e.g., trained on the source domain) and a model decided by a certain index on a particular land-cover class, e.g., NDVI for

## TABLE V
### INTRODUCTION OF INDICES

| | Equation | Description | Threshold |
|---|---|---|---|
| **NDVI**[13] | $NDVI = \dfrac{NIR - RED}{NIR + RED}$ | Extract vegetation using NIR and RED bands | >Otsu's method: tree |
| **NDWI**[42] | $NDWI = \dfrac{GREEN - NIR}{GREEN + NIR}$ | Extract water using NIR and GREEN bands | >Otsu's method: water |
| **MBI-H**[40] | $MBI\text{-}H = DSM - B_{DSM\varepsilon(DSM,e)}$ | Extract building using Morphology Top-Hat Reconstruction on DSM. $\varepsilon$ is the erosion operation and $e$ is the structuring element. $B$ is the Reconstruction operation | >2: building |

the vegetation class, in which X is the input (pixel values or transformed features), Y and $\overline{Y}$ are the random variable representing the output of both models (the supervised model and the index-based prediction model) being either correct (a value of 1) or incorrect (a value of 0), conditioned by the input X. Considering a binary case, the probability of these two predictions being consistent (i.e., $\mathcal{P}(Y = \overline{Y}|X)$), follows:

$$\mathcal{P}(Y = \overline{Y}|X) = \mathcal{P}_{\text{sup}}(Y = 1|X) \cdot \mathcal{P}_{\text{index}}(\overline{Y} = 1|X) + \mathcal{P}_{\text{sup}}(Y = 0|X) \cdot \mathcal{P}_{\text{index}}(\overline{Y} = 0|X) \quad (2)$$

where it assumes independent distribution when the input is processed by these two different models. In the target dataset, $\mathcal{P}_{\text{sup}}(Y|X)$ is not accessible due to the lack of target domain labels for training, while $\mathcal{P}_{\text{index}}(\overline{Y}|X)$ can be regarded as independent of training data. We generally regard the positive prediction $\mathcal{P}_{\text{sup}}(Y = 1|X)$ and $\mathcal{P}_{\text{index}}(\overline{Y} = 1|X)$ are better than random guesses (larger than 0.5); thus $\mathcal{P}_{\text{sup}}(Y = 0|X) \cdot \mathcal{P}_{\text{index}}(\overline{Y} = 0|X)$ can be a relatively small number (i.e., $< 0.5 \times 0.5 = 0.25$). Therefore, by omitting this small number, $\mathcal{P}(Y = \overline{Y}|X)$, i.e., prediction evaluated by using the results of indexed-based results, can be loosely understood as proportional to the supervised performance $\mathcal{P}_{\text{sup}}(Y|X)$:

$$\mathcal{P}_{\text{sup}}(Y|X) \propto \mathcal{P}(Y = \overline{Y}|X) \quad (3)$$

This loosely derived conclusion will be held more correct as the index-based predictor is getting more accurate (i.e., $\mathcal{P}_{\text{index}}(\overline{Y} = 1|X)$ is getting close to 1). It may present as a reasonable predictor when a trained classifier is applied to an unlabeled domain, thus, it can give an estimate of the transferability of that model. This is particularly useful when selecting one among multiple trained models to be applied to a target domain. To validate this idea, we performed the experiments using our datasets. The goal is to validate Equation (3), i.e., identify if the accuracy evaluated by index-based labels positively correlates with the accuracy evaluated by the ground-truth label. Experiments and discussions are explained as follows.

### B. Experimental Analysis

There are many remote sensing indices that are widely used to provide cues for vegetation, water, shadow, buildings, etc. Some of these are based on spectral properties (such as NDVI) and some are based on the spatial textures of images (such as morphological properties of pixels [52]). In this study, we evaluate a few typical ones to understand their predicting powers when used following the concept. These evaluated indices include the NDVI, NDWI, and the Morphology Top-hat Reconstruction on DSM (MBI-H) [44] to identify trees, water, and buildings with thresholds, respectively (see TABLE V.).

After extracting these indices, the index-based labels can be obtained by binarizing these indices using thresholds. All the NDWI and NDVI (by default they are -1 ∼ 1) that are smaller than 0 are set to 0. Then, we determine the thresholds of NDWI and NDVI using Otsu's method [53]. For MBI-H (which is the same as the un-normalized AGL), we set a threshold as 2, i.e., all the pixels that have higher MBI-H are regarded as buildings. The priority of those tree indices is that NDVI is higher than MBI-H and followed by NDWI. It should be noted that when the height information is not available, one can use MBI [44] followed by Otsu thresholding as an alternative.

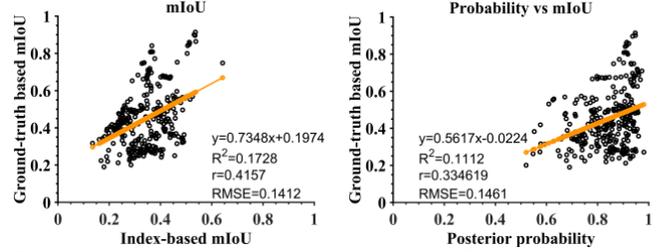

**Fig. 5:** Correlation of (a) index-based mIOU (our approach) and (b) posterior probability from model confidence measures, with the ground-truth mIOU. Details can be found in the text.

With the above-mentioned index-based thresholding, we can derive index-based labels. Herein, we define the model performance evaluated by the index-based labels as index-based mIOU, and that evaluated by the ground-truth labels as ground-truth-based mIOU. When evaluated by indices, the performance of a trained model varies greatly among different target datasets. This is to verify if such a variation is consistent with the transferability of the model on these domains. Therefore, this correlation is drawn in **Fig. 5 (a)**. Moreover, we compare our approach with typical model confidence measures, i.e., averaged posterior probability of model outputs (i.e., Softmax derived class probability), and draw its correlation with the ground-truth based mIOU. Our approach has a higher correlation with the Ground-truth mIOU, evidenced by the higher $R^2$ value (and other statistics in **Fig.5**), which further validate the effectiveness of our approach.

### VI. CONCLUSION

In this work, we perform a comprehensive study on three connected aspects of the model transferability for remote sensing data semantic segmentation. ***First***, we quantitatively study the transferability of learned models when applied to remote sensing datasets for semantic segmentation applications. Unlikely typical transfer learning studies which are designated to develop new learning strategies or merely describe generalization problems in a qualitative fashion, this work evaluates the raw transferability by harnessing four large volume, public/customized remote sensing datasets, to quantitatively understand the varied performance of six trained models (one traditional and five deep learning models) when applied to datasets of different sensors and different geographical locations. ***Second,*** we evaluate three state-of-the-art deep learning-based domain adaptation (DA) methods in terms of their performance in improving the model's transferability (i.e., "closing" the domain gaps), to analyze the so-called adapted transferability. ***Third***, we have proposed and validated a simple, but effective strategy to evaluate the raw transferability of a model without needing any ground-truth data, which can be particularly beneficial to allow automated model selection for model ensemble.

The study yields several general yet not well-reported conclusions: ***First***, we found that when a trained model is applied to a different dataset, the performance can vary and is significantly low (30+% lower). With DA, the performance has a notable improvement but is generally smaller than 10% and thus, there is a large room for improvement. ***Second***, the raw transferability of a model is not specifically tied to the volume of training data, but the quality of the data that can capture "good" distribution, while this aspect remains to be researched. ***Third***, a better-performed model in the S2S test, does not necessarily guarantee a relatively better performance in the S2T test. Therefore, the ability to evaluate the raw transferability to take preventative measures on bad generalization is still very important. ***Fourth***, we found that height information can be a valid source to improve the raw transferability of models, due to its explicit geometric information. For example, when using the height information, the most improvement is around 12% when using a shallow classifier (RF), while less improvement is found with more complex models, and less room for improvement by DA. ***Fifth***, the performance of the DA approaches varies with the base model, while among the three compared DA approaches, we found that the ScaleAware [13] approach consistently achieves the best performance (regardless of the used base models). More detailed and specific observations can be found in Section IV.

Additionally, we have validated a simple idea of the transferability of a trained model to a target domain (with no labels), which uses index-based labels (as pseudo-ground-truth) to evaluate the performance, called index-based mIOU measure. We found this measure, when correlated with real ground-truth-based mIOU, demonstrates a higher correlation (R-square value) than that of the typical posterior probability of the classifiers. Since this measure is computed by directly using the prediction results, we consider it to be more robust, and can be extremely helpful when model ranking in transfer learning and models ensemble.

Considering the demands and scarcity of labeled VHR datasets, this study has gathered reasonably rich data for processing and analysis, to investigate model transferability with respect to data, models, and DA approaches. The transferability study in this work builds a solid foundation for selecting base models and building DA approaches upon. Although, the studied models are based on the application of semantic segmentation, the observations and conclusions may be generally applicable to other applications. Our experiments and the above conclusions suggest that the development of more generalized and large models, should not only based on the DA approaches, but also based on the richness of training data, the base model, and the spectral or physics-based domain-specific indicators (indices and height information). This work focuses on VHR datasets, and future studies may include the analysis of datasets with various resolutions.

### Acknowledgment

We thank the organizers of the 2019 Data Fusion Contest and DigitalGlobe for providing the JAX and OMA datasets.